\documentclass[sigconf]{acmart}

\copyrightyear{2022}
\acmYear{2022}
\setcopyright{acmlicensed}\acmConference[MM '22]{Proceedings of the 30th ACM International Conference on Multimedia}{October 10--14, 2022}{Lisboa, Portugal}
\acmBooktitle{Proceedings of the 30th ACM International Conference on Multimedia (MM '22), October 10--14, 2022, Lisboa, Portugal}
\acmPrice{15.00}

\usepackage{amsmath}

\usepackage{amssymb}
\usepackage{multirow}
\usepackage{mathrsfs}
\usepackage{array}
\usepackage{color}
\usepackage{ifpdf}
\usepackage{algorithmic}
\usepackage{algorithm}
\usepackage{balance}
\begin{document}
\title{Exploring Negatives in Contrastive Learning for Unpaired Image-to-Image Translation}


\author{Yupei Lin}
\authornote{Both authors contributed equally to this research.}
\affiliation{%
  \institution{Guangdong University of Technology}
  \city{Guangzhou}
  \country{China}}
 \email{yupeilin2388@gmail.com}

\author{Sen Zhang}
\authornotemark[1]
\affiliation{%
  \institution{The University of Sydney}
  \city{Sydney}
  \country{Australia}}
\email{szha2609@uni.sydney.edu.au}

\author{Tianshui Chen}
\affiliation{%
  \institution{Guangdong University of Technology}
  \city{Guangzhou}
  \country{China}}
\email{tianshuichen@gmail.com}

\author{Yongyi Lu}
\affiliation{%
  \institution{Guangdong University of Technology}
  \city{Guangzhou}
  \country{China}}
\email{yylu1989@gmail.com}

\author{Guangping Li}
\affiliation{%
  \institution{Guangdong University of Technology}
  \city{Guangzhou}
  \country{China}}
\email{gpli@gdut.edu.cn}

\author{Yukai Shi}
\authornote{Corresponding author: Yukai Shi}
\affiliation{%
  \institution{Guangdong University of Technology}
  \city{Guangzhou}
  \country{China}}
\email{ykshi@gdut.edu.cn}
\renewcommand{\shortauthors}{Yupei Lin , et al.}
\begin{abstract}
Unpaired image-to-image translation aims to find a mapping between the source domain and the target domain. To alleviate the problem of the lack of supervised labels for the source images, cycle-consistency based methods have been proposed for image structure preservation by assuming a reversible relationship between unpaired images. However, this assumption only uses limited correspondence between image pairs. Recently, contrastive learning (CL) has been used to further investigate the image correspondence in unpaired image translation by using patch-based positive/negative learning. Patch-based contrastive routines obtain the positives by self-similarity computation and recognize the rest patches as negatives. This flexible learning paradigm obtains auxiliary contextualized information at a low cost. As the negatives own an impressive sample number, with curiosity, we make an investigation based on a question: are all negatives necessary for feature contrastive learning? Unlike previous CL approaches that use negatives as much as possible, in this paper, we study the negatives from an information-theoretic perspective and introduce a new negative Pruning technology for Unpaired image-to-image Translation (PUT) by sparsifying and ranking the patches. The proposed algorithm is efficient, flexible and enables the model to learn essential information between corresponding patches stably. By putting quality over quantity, only a few negative patches are required to achieve better results. Lastly, we validate the superiority, stability, and versatility of our model through comparative experiments.
\end{abstract}



\maketitle
\section{Introduction}

The image-to-image translation task aims to match the style of images from a source domain into a target domain, while retaining the original structure of the source images~\cite{cai2021frequency,kwon2021diagonal,wang2021transferi2i}. For current unpaired image-to-image translation datasets, the generative adversarial nets (GAN)~\cite{goodfellow2014generative,richardson2021encoding,zhou2021cocosnet,zhan2021unbalanced,li2021image} are able to generate images that match the style, but the adversarial loss suffers from the collapse of the structure. This problem may lead to translated images with poor qualities. Thus, improving the quality of image translation with consistent structure remains a very challenging task.

To alleviate this problem, a family of methods based on cycle-consistency learning~\cite{zhu2017unpaired,cai2021frequency,hadji2021representation,ge2021disentangled,kim2019u} have been proposed. Cycle-consistency assumes that there exists a reversible relationship between the source image and the target image. Taking advantage of this relationship, the generative adversarial network can preserve the consistency of the image structure. Nevertheless, this assumption has a limited flexibility, since the cyclic loss solely emposes strict alignment on the structures without exploiting other knowledge.

\begin{figure*}[h]
\centering
\includegraphics[width=0.98\linewidth]{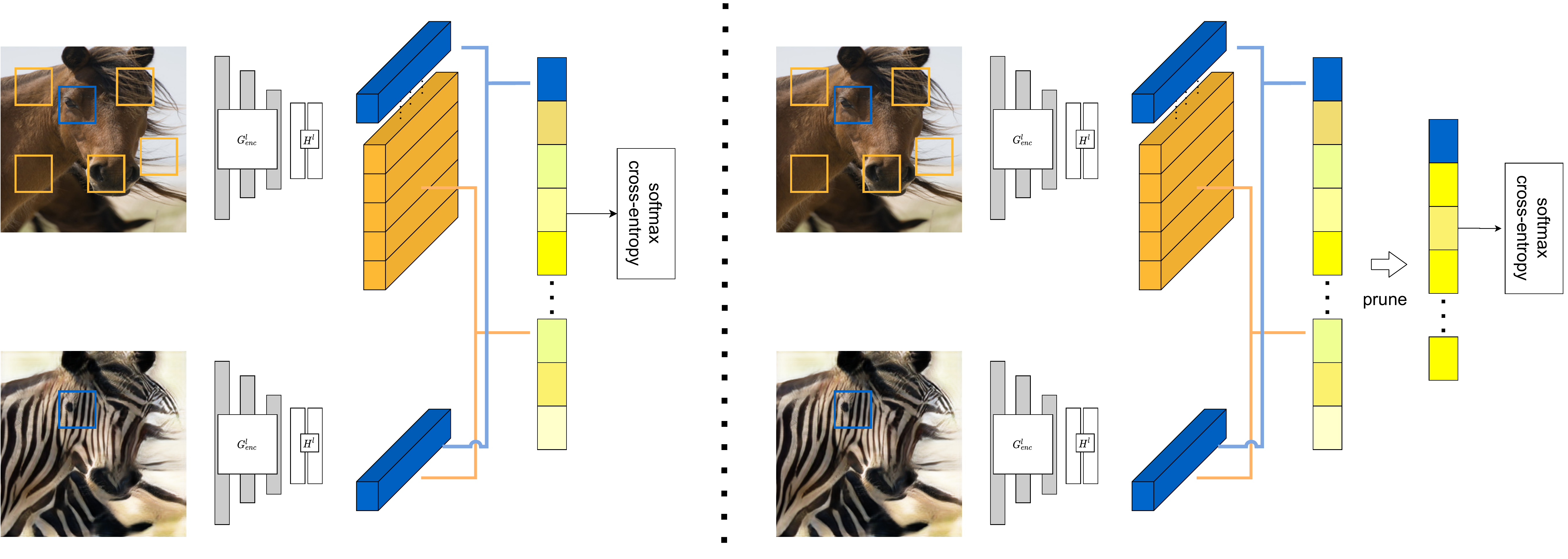}
\vspace{-3mm}
\caption{Left: The PatchNCE loss~\cite{park2020contrastive}. Right: the proposed RankNCE loss. A typical patch-wise contrastive paradigms usually maximize the mutual information on all the samples, regardless of positive or negative. In comparison, RankNCE selects negative samples with good qualities based on their contribution to the mutual information (MI).}
\vspace{-3mm}
\label{fig:PCL}
\end{figure*}

Recently, CUT~\cite{park2020contrastive} refreshes the cycle-consistent paradigm by incorporating contrastive learning (CL) into unpaired image-to-image (I2I) translation. By using CL, the generative model is able to use auxiliary knowledge to complement the cycle-consistency. As shown in Fig.~\ref{fig:PCL}, given a query sample, CUT chooses to use an assigned patch (i.e., in the same location) as the positive target for cycle-consistent learning. Meanwhile, CUT cleverly uses non-local image structures as the negative samples. This flexible learning paradigm achieves high-quality image translation without any auxiliary supervision on the training image. However, CUT uses all non-local patches as negative samples, which leads to an obvious problem, i.e., a large number and a potentially high quality diversity of the negative samples. As a result, NEGCUT~\cite{wang2021instance} points out that such unbalanced distribution of negative samples may increase the difficulty of learning, since the information of more relevant negative samples may be overwhelmed by the others. To address this phenomenon, NEGCUT develops a generator to produce fake negative samples. However, NEGCUT still suffers from drawbacks such as the additional computation cost, conflicting quality of negatives, and the increased training difficulty. These problems motivate us to develop an efficient yet advantageous negative selection strategy to learn the essential correspondence between samples.


As suggested in~\cite{wang2020understanding}, in CL, a uniformity property is usually desired for the features to maintain as much information as possible, which means the high-dimensional features are expected to uniformly distributed on a hypersphere. In this sense, introducing too many negative pairs with diverse qualities in contrastive representation learning would go against this uniformity property. Inspired by RankSRGAN~\cite{zhang2019ranksrgan}, we propose a ranking technology for negative samples, by making the following hypothesis: fewer samples with good quality are enough for CL. Thus, the proposed method attempts to investigate a selective strategy on negative samples. As shown in Fig.~\ref{fig:Visual}, to exhibit a discriminative CL, only a few negative patches that contain the most valuable information are carefully preserved. Specifically, we construct negative pruning technology with three steps: i.e., the similarity score computation, pruning, and ranking phrases. In addition, we re-examine PUT in the information-theoretic language to strengthen the interpretability of our method. Overall, the proposed negative pruning technology encourages the generative model to learn essential features, reduces the difficulty of learning, and helps to stabilize training. Our main contributions are:

\begin{figure*}[h]
\centering
\includegraphics[width=0.95\linewidth]{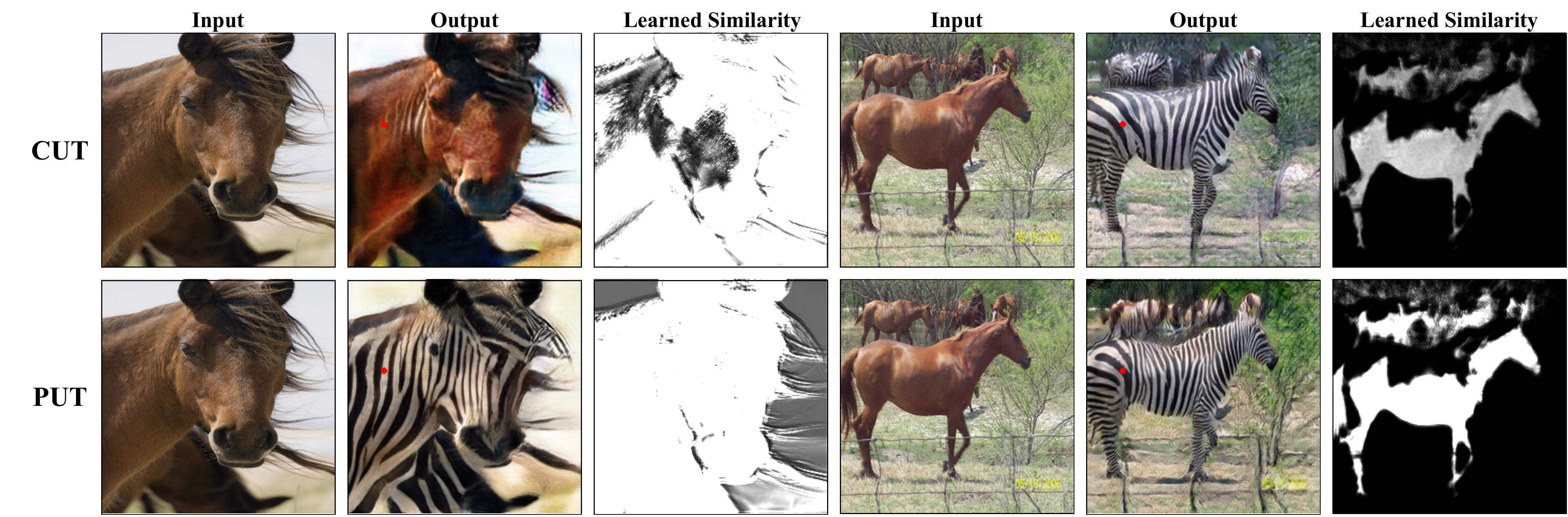}
\vspace{-3mm}
\caption{Visualization of the learned similarity by the feature extractor. Given an input and output image, we extract the features of these images through a feature extractor. We compute the learned similarities between the feature vectors of $[(v,v^-_1),...,(v,v^-_N)]$ by using exp($v \cdot v^- / \tau$). Specifically, \( v \) is a query element (the highlighted red dot in the output) and $[v^-_1,...,v^-_N]$ are all the candidate patches in the input. In contrast to CUT, the feature extractor of our model learns the cross-domain correspondence with a better saliency effect.}
\label{fig:Visual}
\vspace{-4mm}
\end{figure*}
\begin{itemize}

\item Unlike previous contrastive learning methods that use all the negative samples, we investigate strategies to utilize fewer but better negative samples. Specifically, we develop a negative pruning technology, which consistently shows better results with fewer negative samples.

\item To provide further insights and interpretability of our method, we re-examine the image-to-image translation problem from an information-theoretic perspective, by investigating the mutual information between the contrastive pairs. 

\item In the experiments, we show that PUT brings better superiority and stability. And each sub-component in PUT is carefully analyzed to verify its effectiveness.

\end{itemize}

The rest of the paper is organised as follows: In Section~\ref{sec:related}, we present related work, including current methods related to image-to-image translation and contrastive learning. In Section~\ref{sec:method}, we introduce our methods, including preliminaries and each component of PUT. In Section~\ref{sec:experiment}, we present the experimental procedure, the comparison with existing methods, and the results of the ablation experiments. Finally, conclusion and limitations are given in Section~\ref{sec:conclusion}. 

\section{Related work}
\label{sec:related}

\textbf{Image-to-Image (I2I) translation.} 
Image-to-image translation aims to learn the style of images from a source domain into a target domain~\cite{shao2021spatchgan,jia2021semantically,wang2021instance,baek2021rethinking,xie2021unaligned}, which can be classified into two groups: the paired setting\cite{isola2017image,isola2017image,wang2018high,park2019semantic} (supervised) and an unpaired setting (unsupervised). Paired setting means the training set is supervised, each image from source domain has a corresponding label from target domain. Pix2Pix~\cite{isola2017image}, Pix2PixHD~\cite{wang2018high} and SPADE~\cite{park2019semantic} use adversarial loss~\cite{goodfellow2014generative} in paired training data to train their model. Unlike paired settings, instances of the unpaired setting have no corresponding label from the target domain. To enable to model training with an unsupervised condition, the cycle-consistency has become a popular scheme, which learns a reversible projection from the target domain back to the source domain. For example, UNIT~\cite{liu2017unsupervised}, DualGAN~\cite{yi2017dualgan} and MUNIT~\cite{huang2018multimodal} train cross-domain GANs with nested cycle-consistent losses. However, this paradigm is often too restrictive to obtain sufficient context. To this end, many approaches attempt to design a unidirectional translation to eliminate the cycle-consistent constraint~\cite{pang2021image}. DistanceGAN~\cite{benaim2017one} proposes a distance constraint that allows unsupervised domain mapping to be one-sided. GC-GAN~\cite{fu2019geometry} enforces geometry consistency as a constraint for unsupervised domain mapping. In recent method, CUT~\cite{park2020contrastive} introduces contrast learning in unpaired settings. CUT~\cite{park2020contrastive} introduces contrast learning to obtain auxiliary negative labels. However, these methods still suffer from a potentially large number of negative samples with diverse qualities.

\textbf{Contrastive representation learning.} 
Contrastive learning is a powerful scheme for self-supervised representation learning. It has achieved good results in the field of unsupervised representational learning~\cite{hjelm2018learning,he2020momentum,van2018representation,chen2020simple,henaff2020data}. Contrastive learning uses a noisy contrastive estimation framework~\cite{gutmann2010noise} to maximize the mutual information of corresponding patches between images by comparing positive pairs with negative pairs. PatchNCE~\cite{park2020contrastive} proposes patch-based contrastive learning, which uses a noise-contrastive estimation framework in unpaired I2I translation tasks by learning the correspondence between the patches of the input image and the corresponding generated image patches. Excellent results are achieved and the recent methods also obtained better performance by utilizing the idea of patch-wise learning. DCLGan~\cite{han2021dual} uses contrastive loss with a dual-way setting, F-LSeSim\cite{zheng2021spatially} proposes a systematic way to obtain a general spatially-correlative map. In parallel to these various designed methods, we mainly investigate negatives sample selection strategies by introducing the idea of ranking in the contrastive learning, which consistently shows better results with fewer negative samples.

\section{Method}
\label{sec:method}
\begin{figure*}[h]
\centering
\includegraphics[width=0.97\linewidth]{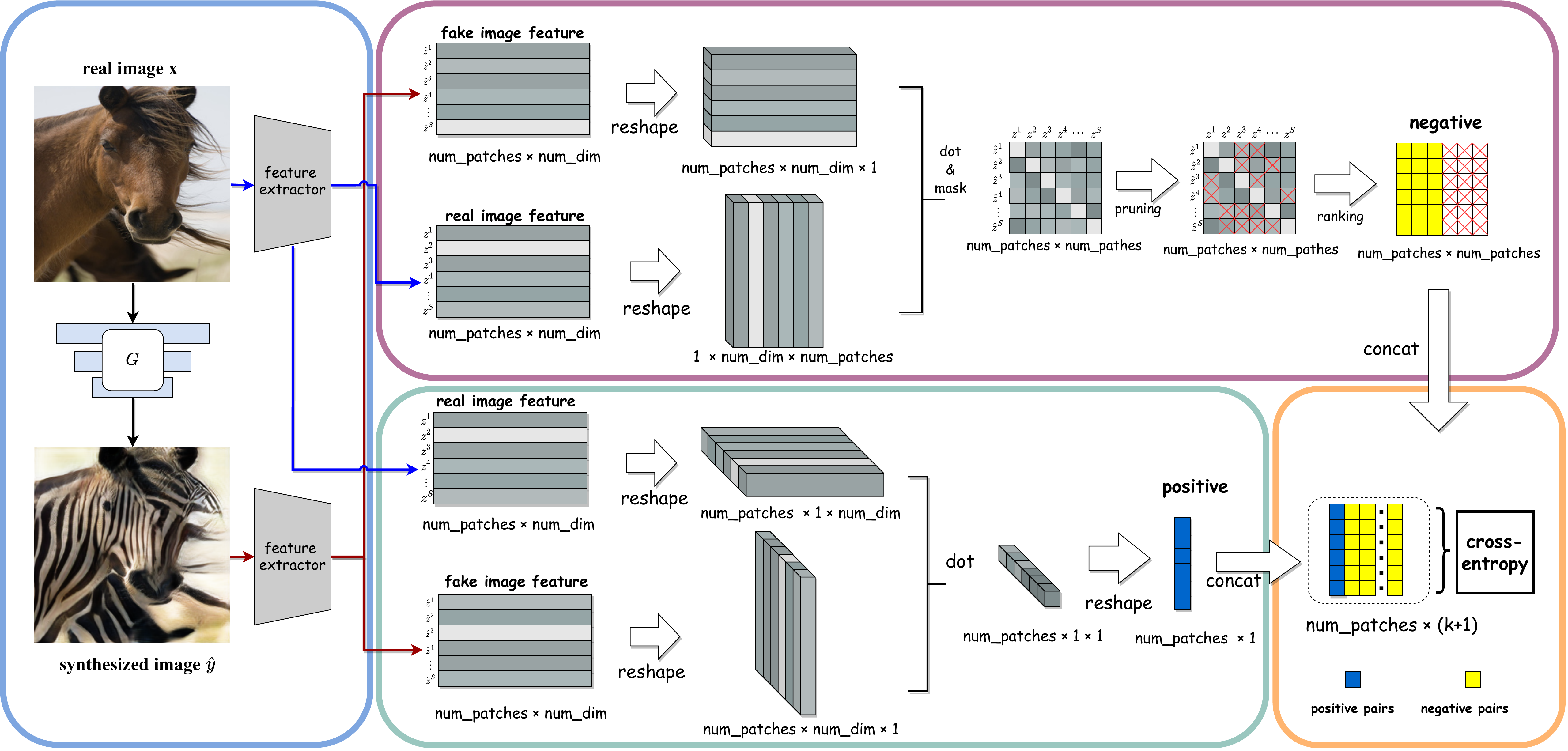}
\vspace{-3mm}
\caption{The proposed RankNCE loss. The real images x and the synthesized image \(\hat{y}\) are firstly transformed into feature vectors using the generator \(G\). The synergistic relationship is computed by applying dot-product on homogeneous features. The contrastive relationship is obtained by multiplying determinantal features of fake/real images. Our proposed RankNCE differs from existing methods by pruning and ranking rough negatives.}
\vspace{-3mm}
\label{fig:patchnce}
\end{figure*}

In Fig.~\ref{fig:patchnce}, we show the architecture of PUT, which includes two types of losses (i.e., adversarial loss and RankNCE loss). In particular, RankNCE is formed by pruning and ranking operations. Firstly, the input image \(x\) is mapped by the generator \(G:x\to y\) to generate the image \(\hat{y}\). Secondly, the feature similarity between corresponding patches and non-corresponding patches is calculated at each layer to obtain mutual information (MI). Finally, top-k patches with the highest similarity are selected by using cross-entropy loss for sorting. 
\subsection{Preliminaries}
\textbf{Unpaired I2I Translation.} Given domain $ X \subset  \mathcal{R}^{H \times W \times C}$, we hope that the image will be as similar as possible to the image \(Y \subset  \mathcal{R}^{H \times W \times C}\) with target domain after translation. Given a dataset of unpaired instances \(\lbrace x \in X \rbrace\) and \( \lbrace y \in Y \rbrace \), the \(G:x \to  y\) mapping is completed with a generative network.

To realize the mapping \(G:X \to  Y\), we use the adversarial loss to encourage the generative model to produce images that are visually similar to the target domain image. The expression for the adversarial loss is expressed as follows:
\begin{equation}
  \mathcal{L}_{GAN}(G,D_Y,X,Y) = E_{y\sim Y}logD(y)+E_{x\sim X}log(1-D(G(x))) 
\end{equation}
 
\textbf{PatchNCE.}
According to contrastive predictive coding~\cite{park2020contrastive,van2018representation}, a good translated image should have high mutual information with corresponding patches of the input image. To maximize mutual information, PatchNCE~\cite{park2020contrastive} employs the contrastive feature learning on internal patches. This loss simply applies a noisy contrastive estimation framework by introducing positive and negative pairs within the image. Specifically, a corresponding positive patch is generated by applying the tokenization step to an input image. And the negative pairs are the patch of the non-corresponding tokens. This method correlates input and output information in a non-local fashion, and the generative model learns rich information from the synergistic/contrastive relationship. Formally, let \(v  \in R^C\)  denotes the C-dimensional feature vector of the `query' patch, \( v^+ \in R^C\) denotes the feature of the positive patch corresponding to the query patch and \(v^- \in R^{C\times N} \) denotes the features of N negative patches. The contrastive loss is defined as an (N+1)-way classification problem, and is expressed as:

\begin{equation}
\ell(v,v^+,v^-)=-log(\dfrac{exp(v \cdot{v^+})/\tau}{exp(v \cdot{v^+})/\tau+\sum^N_{i=1}{exp}(v \cdot{v^-_i})/\tau}), \label{eq:cl_loss},
\end{equation}
where \( \tau \) indicates the temperature coefficient, \(N\) is the number of negative patches. In Eq.~\ref{eq:cl_loss}, \( \tau=0.07 \) and \(N = 255\).

\subsection{RankNCE} Based on  Eq.~\ref{eq:cl_loss}, we review negative samples by observing that the negatives have a large sample number and MI variance. By putting quality over quantity, we attempt to sort out valuable patches to realize a discriminative representation learning under the contrastive paradigm. We therefore modify Eq.~\ref{eq:cl_loss} by introducing the idea of rank. The RankNCE loss is expressed as:
\begin{equation}
 \ell_{rank}(v,v^+,v^-)=-log(\dfrac{exp(v \cdot{v^{+}})/\tau}{exp(v \cdot{v^+})/\tau+\sum^K_{i=1}{exp}(r(v \cdot v^-_i,K))/\tau})
 \label{eq:rank}
 \end{equation}

\begin{figure*}[h]
\centering
\includegraphics[width=0.89\linewidth]{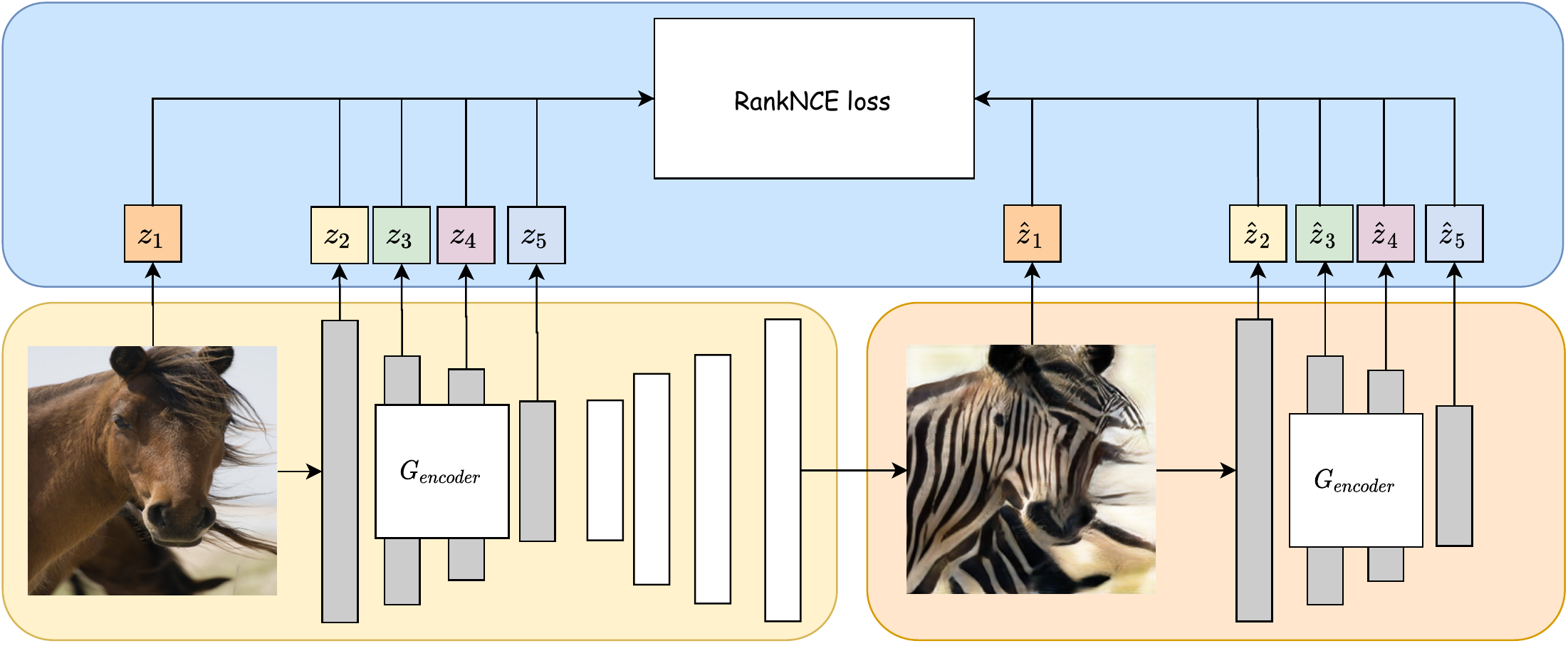}
\vspace{-3mm}
\caption{A illustration of multi-layer RankNCE loss.}
\label{fig:multi_rank}
\vspace{-3mm}
\end{figure*}

where \(v\cdot v^-\) is the similarity between query patch and negative patch, k is an empirical value for valuable negative sample preserving. And \(r(v \cdot v^-_n,K)\) denotes the ranking operator to sort out top-K most similar patches, where $K \ll N$. Especially, the ranking operator contains three-step: \emph{similarity score computation, pruning and ranking.}

\emph{Similarity score computation.} Inspired by InfoNCE~\cite{van2018representation} and PatchNCE~\cite{park2020contrastive}, we first obtain the similary scores between the real image and fake images by calculating the dot-product similarities. As shown in Fig.~\ref{fig:patchnce}, the extracted features of real image and fake image features are $\lbrace z^1,..,z^S \rbrace \subset Z$ and $\lbrace \hat{z}^1,..,\hat{z}^S \rbrace \subset \hat{Z}$, respectively. We obtain the similarity score matrix $C_{sim}$ as:
\begin{equation}
\begin{split}
C_{sim}&= \hat{Z}^T Z, \\
&= [\hat{z}^1,..,\hat{z}^S]^T [z^1,..,z^S], \\
&\overset{mask}{=}\begin{bmatrix}
 0&  \hat{z}^1z^2 & ... &  \hat{z}^1z^S \\
 \hat{z}^2z^1& 0 & ... & \hat{z}^2z^S  \\
 ...& ... & ... & ... \\
 \hat{z}^Sz^1& \hat{z}^Sz^2 & ... & 0 \\
\end{bmatrix},\\
\end{split}
\end{equation}
where $\overset{mask}{=}$ means the diagonal elements are filled with zero.

\emph{Pruning.} To remove meaningless negative samples that were involved in discriminative learning, we distill $C_{sim}$ by sparsifying elements according to $MI(v||v^-)$. In other words, the local samples, which contribute less to the mutual information between the anchor and its negative samples, are restricted from participating in CL. 

\emph{Ranking.} A ranking operator is then applied in $C_{sim}$ to preserve the most valuable elements. As shown in Fig.~\ref{fig:patchnce}, the top-K patches with the highest similarity scores are sorted out according to the y-axis, and the remaining negatives are reset to zero. Though we have used ranking operation, the preliminary pruning scheme is necessary, as the sort operation still has an opportunity to retain negative pairs that contribute less to $MI(v||v^-)$.

Finally, the reserved elements in $C_{sim}$ are utilized as negative samples. As shown in Fig.~\ref{fig:patchnce}, the positive/negative samples are collected for CL with Eq.~\ref{eq:rank}. To further demonstrate the difference between PatchNCE and RankNCE, we compare the learned similarities of each method in Fig.~\ref{fig:Visual}. Intuitively, PUT leans the cross-domain correspondence with a better saliency effect, which justify our strategy is helpful to discriminative representation learning. To provide more insights and interpretability of PUT, we further investigate the MI between the synergistic/contrastive patches from an information-theoretic perspective.

\subsection{Information-theoretic Perspective}
Intuitively, from the information-theoretic perspective, a good CL approach should maximize the MI between $v$ and $v^+$, while minimizing the MI between $v$ and $v^-$ at the same time. It has been shown that minimizing the contrastive loss in Eq.~\ref{eq:cl_loss} can be regarded as maximizing a variational lower bound of the mutual information $MI(v||v^+)$ between the features of the query patch and its positive pair~\cite{poole2019variational}:
\begin{equation}
    MI(v||v^+) \geq E[log\frac{exp(v\cdot v^+)/\tau}{exp(v\cdot v^+)/\tau + \sum_{n=1}^N exp(v\cdot  v^-_n)/\tau}]. \label{eq:MI_0}
\end{equation}
Of note is that the lower bound of $MI(v||v^+)$ in Eq.~\ref{eq:MI_0} makes the underlying assumption that $v^+$ and $v$ are both independent from $v^-$, which will not necessarily hold in practice. Instead, when the independence assumption is violated, we have the following lower bound for the mutual information $MI(v||v^+,v^-)$ between the features of the query patch and its positive/negative pairs~\cite{poole2019variational}:
\begin{equation}
    MI(v||v^+,v^-)\geq E[\frac{1}{N+1}\sum_{i=0}^{N} log\frac{exp(v_i\tilde{v_i}^+)/\tau}{\frac{1}{N+1}\sum_{k=0}^N exp(v_k\tilde{v_i}^+)/\tau}], \label{eq:MI_1}
\end{equation}
where $v_0=v^+$ and $v_i=v_i^-$ for $i\geq 1$. $\tilde{v_i}^+$ denotes the feature of a corresponding positive patch w.r.t. $v_i$, where $\tilde{v_0}^+=v$. Since for each training step we do not have the information of $\{\tilde{v_i}^+\}_{i=1}^N$ for $v^-$, we can use $-\ell(v,v^+,v^-)$ as a sample estimate for the expectation in Eq.\ref{eq:MI_1}, by assuming a symmetric relationship among all patches: i.e. $E[log\frac{exp(v_i\tilde{v_i}^+)/\tau}{\frac{1}{N+1}\sum_{k=0}^N exp(v_k\tilde{v_i}^+)/\tau}] = E[log\frac{exp(v_j\tilde{v_j}^+)/\tau}{\frac{1}{N+1}\sum_{k=0}^N exp(v_k\tilde{v_j}^+)/\tau}]$, $\forall i\neq j$. Therefore, minimizing the PatchNCE loss is essentially equivalent to maximizing the variational bound of $MI(v||v^+,v^-)$.

While $MI(v||v^+)$ could potentially be increased by maximizing this lower bound of $MI(v||v^+,v^-)$, however, a direct use of the PatchNCE loss does not consider $MI(v||v^-)$ explicitly, and a unbounded $MI(v||v^-)$ may lead to sub-optimal results. \emph{Inspired by the observation that not all $\{v^-_n\}_{n=1}^N$ contribute equally to $MI(v||v^-)$ and thus the CL process}, we propose to put more focuses on negative features that potentially 
lead to a large $MI(v||v^-)$, which indicates these negative features are not well learnt. 

\begin{figure*}[h]
\includegraphics[width=0.98\linewidth]{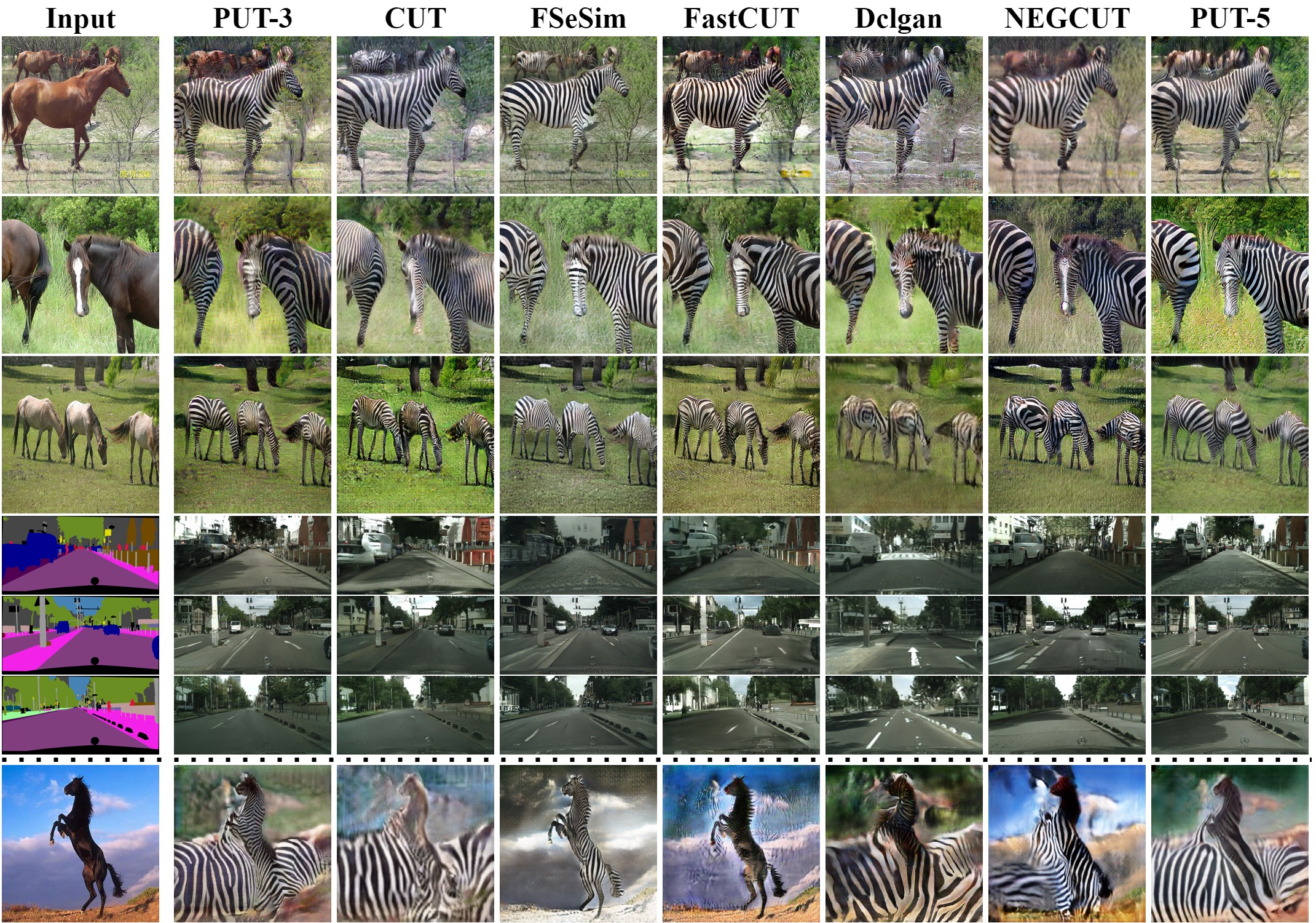}
\vspace{-3mm}
\caption{\textbf{Qualitative results.} We compare the performance of our method with state-of-the-art methods on the Horse$\rightarrow$Zebra and CityScapes datasets, where PUT-3 and PUT-5 synthesize realistic textures with consistent structure and brightness. In particular, a typical failure case is visualized in the last row to address the limitation.}
\label{fig:result}
\vspace{-3mm}
\end{figure*}

Specially, we formulate the conditional probabilities as:
\begin{align}
    p(v|v^+) &= p(v)\frac{exp(v\cdot v^+)/\tau}{exp(v\cdot v^+)/\tau + \sum_{n=1}^N exp(v\cdot v^-_n)/\tau}, \label{eq:prob_0} \\
    p(v|v^-_n) &= p(v)\frac{exp(v\cdot v^-_n)/\tau}{exp(v\cdot v^+)/\tau + \sum_{i=1}^N exp(v\cdot v^-_i)/\tau},\  1\leq n \leq N. 
    \label{eq:prob_1}
\end{align}
We aim at replace $N$ in Eq.~\ref{eq:prob_1} with a smaller value $K$ by selecting top-ranked negative features. To this end, for the mutual information $MI(v||v^-)$, let $H(\cdot)$ denotes the entropy, then we have:
\begin{align}
    MI(v||v^-) &= H(v) - H(v|v^-) \\
    &= H(v) + E_{v^-}[E_{v|v^-}[log(p(v|v^-))]].
\end{align}
Given $v$ and $v^-$, we can use $p(v|v^-_n)log(p|v^-_n)$ as a sample estimate of $E_{v|v^-}[log(p(v|v^-))]$ for each $v^-_n, n=1,..,N$. Therefore, a larger $p(v|v^-_n)$ will also lead to a larger $MI(v||v^-)$. Since $p(v|v^-_n)$ is monotonically increasing w.r.t. the similarity score $v\cdot v^-_n$ between $v$ and $v^-_n$, we thus use $v\cdot v^-_n$ as the measure of the contribution of $v^-_n$ to $MI(v||v^-_n)$, and construct the NCE loss by selecting the top-ranked negative features so as to focus on the potential negative features that may lead to the MI explosion.

\subsection{Multi-layer Contrastive Rank }

As shown in Fig.~\ref{fig:multi_rank}, a multi-layer contrastive RankNCE is used for a multi-scale training. We formulate the new multi-level loss function based on Eq.~\ref{eq:prob_0} and~\ref{eq:prob_1} is expressed by:
\begin{equation}
\mathcal{L}_{RankNCE}(G,X) = E_{x\sim X}\sum\limits_{l=1}^{L}\sum\limits_{s=1}^{S_l}\lbrace p(\hat{z}_l^s|z_l^s) + p(\hat{z}_l^s|z_l^{S\backslash s}) \rbrace_L.
\end{equation}
We use \(G \) to generate the feature vectors \(\lbrace z_{{l}}\rbrace_L = \lbrace{(G^{{l}}(x))}\rbrace_L \), where \(G^{{l}}\) denotes the output of the \(l\)-th layer in  generator, layers \(l \in \lbrace1,2,...,L\rbrace\). We refer to the corresponding patch feature as \(z^s_l \in \mathbb{R}^{C_l} \), and the none-corresponding patch feature called \(z_{l}^{S\backslash s} \in \mathbb{R}^{(S_l-1)\times C_l}\), where \(C_l\) is the number of channels in the \(l\)-th layer, \(S_l \) denotes the number of spatial positions in the \(l\)-th layer, and \( s\in \lbrace 1,...,S_l\rbrace\). Since the corresponding patches in the real and synthesized images should share the same correspondence. Thus the RankNCE is used for identity-preserving. The overall loss is expressed as follows:
\begin{equation}
\begin{aligned}
\mathcal{L}(G,D,X,Y)=\lambda_{GAN}(\mathcal{L}_{GAN}(G,D,X,Y)+\lambda_X\mathcal{L}_{RankNCE}(G,X)  \\ {+\lambda_Y\mathcal{L}_{RankNCE}(G,Y)}.
\end{aligned}
\end{equation}
We set the \(\lambda_{X} = 1,\lambda_{y} = \)1. Compared to the previous method, our RankNCE presents a pruning strategy to select $v^-$ for further optimization based on their contributions to $MI(v||v^-)$.

\section{Experiments}
\label{sec:experiment}

\subsection{Datasets and Evaluation metrics}
We evaluate our proposed method and baselines on two different datasets, i.e., the \textbf{CityScapes}~\cite{cordts2016cityscapes} and \textbf{Horse$\rightarrow$Zebra}~\cite{zhu2017unpaired} datasets. The Fréchet Inception Distance\cite{heusel2017gans} (FID) is used to measure the visual quality of generated images. For the Cityscapes dataset, we obtain the segmentation mask of each method by a pre-trained DRN-32~\cite{yu2017dilated} model. Then, we use mean average precision (mAp), pixel accuracy (pixAcc) and average class accuracy (classAcc) for evaluation. We employ competitive methods for comparison, including CUT~\cite{park2020contrastive}, Fast CUT~\cite{park2020contrastive}, FSeSim~\cite{zheng2021spatially}, DCLGan~\cite{han2021dual} and NEGCUT~\cite{wang2021instance}.



\begin{figure*}[h]
\centering
\includegraphics[width=0.93\linewidth]{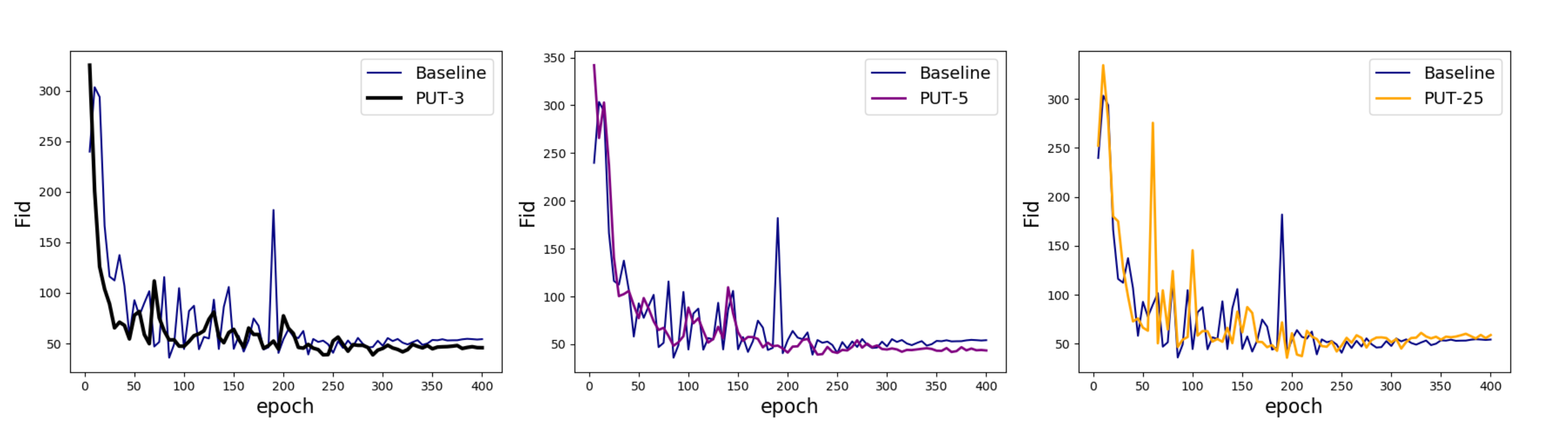}
\vspace{-3mm}
\caption{Training stability on Horse$\rightarrow$Zebra. `PUT-K' indicates the K number of top-ranked negative patches used in RankNCE. Better stability and superior performance are achieved by PUT with fewer negative patches. }
\vspace{-3mm}
\label{fig:Stable}
\end{figure*}

\begin{figure*}[h]
\centering
\includegraphics[width=0.93\linewidth]{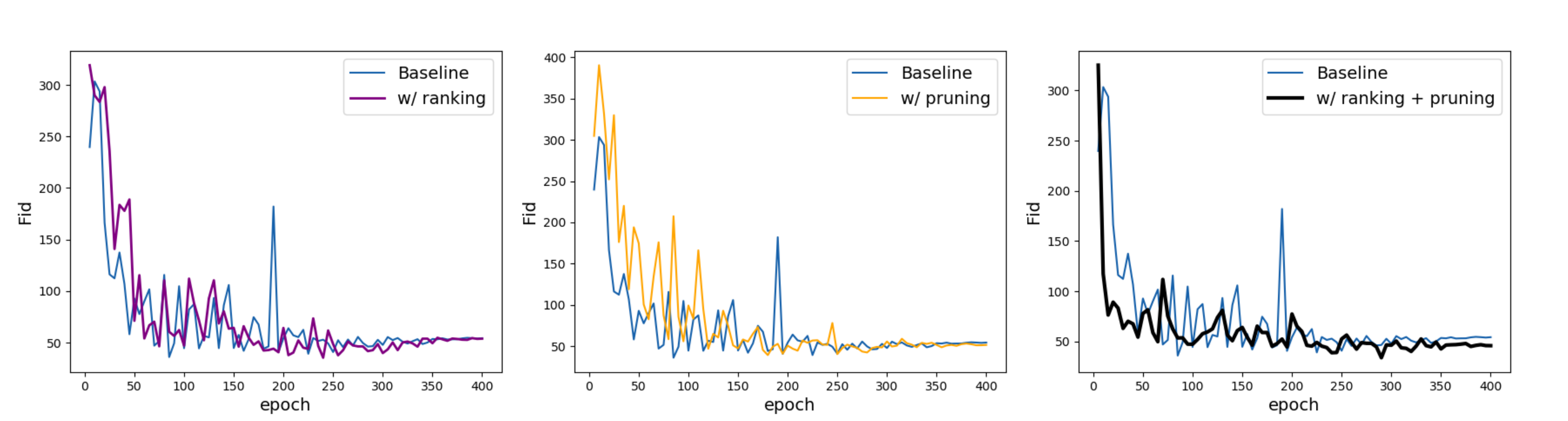}
\vspace{-3mm}
\caption{Ablations of RanKNCE on Horse$\rightarrow$Zebra. We use `sparsifying' and 'ranking' sub-modules separately for ablation study. }
\vspace{-3mm}
\label{fig:ablation_put}
\end{figure*}

\begin{figure*}[t]
\centering
\includegraphics[width=0.93\linewidth]{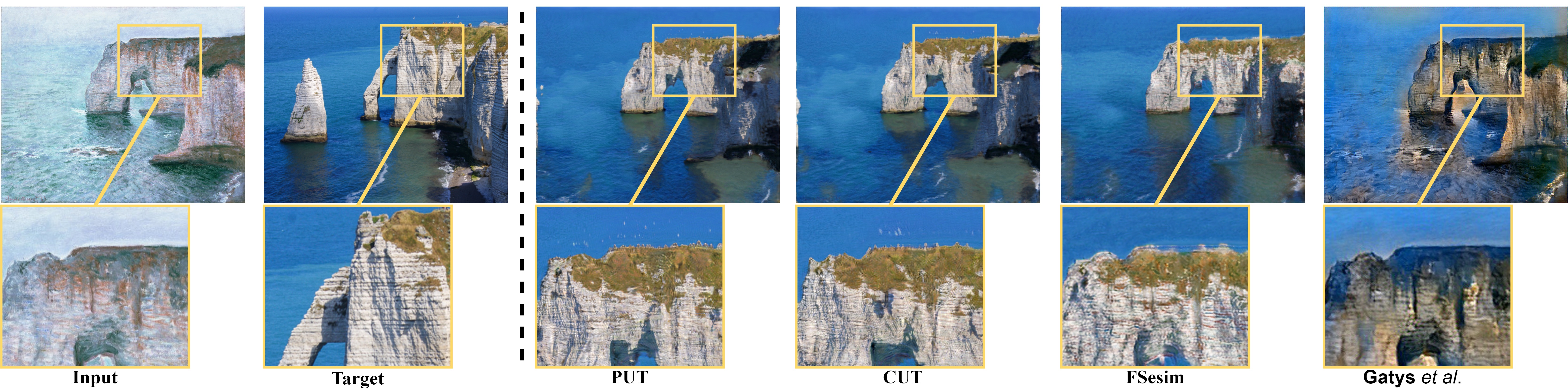}
\vspace{-3mm}
\caption{High-resolution effects. Our method successfully captures the style of the target image while preserving the original structure. Zooming up for a better view.}
\label{fig:hr_effect}
\vspace{-3mm}
\end{figure*}

\begin{figure}[h]
\centering
\includegraphics[width=0.9\linewidth]{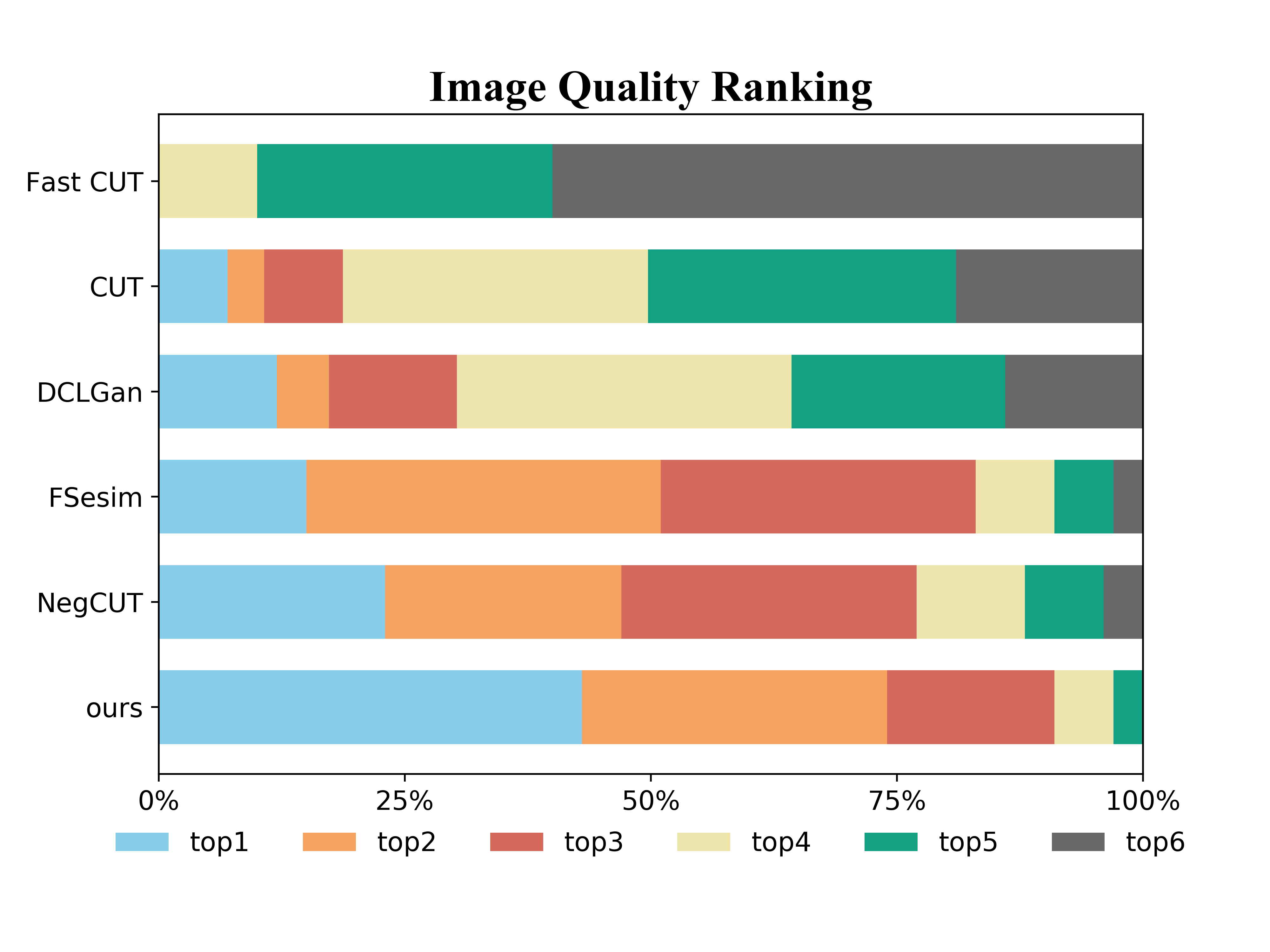}
\vspace{-2mm}
\caption{User study results. We conducted an in-depth quality study by ranking the user's mean opinion score(MOS). }
\vspace{-5mm}
\label{fig:ABtest}
\end{figure}

\subsection{Implementation Details}
We train the model for 400 epochs, and the training batchsize is 4. The learning rate is set to 0.0002, and the learning rate starts to decay linearly after half of the total epoch. Depending on the RankNCE, the number of top-ranked negative samples is set from 3 to 25 for an ablation study.

\subsection{Comparison} 
In Tab.~\ref{tab:my-table}, we show a comparison of the quantitative results. Compared with current state-of-the-art methods, our approach achieves better results with competitive efficiency. For example, PUT surpasses NEGCUT with a 6.01 FID score. Compared with CUT, our model obtains a \emph{significant improvement with similar inference speed,} that justifies the proposed algorithm is solid. As shown in Tab.~\ref{tab:my-table}, the semantic segmentation results on CityScapes also indicate that PUT obtains a better correspondence between output and input. As shown in Fig.~\ref{fig:result}, baseline methods produce blurry textures with poor brightness, while PUT still synthesizes realistic textures with consistent structure and brightness toward the input image.
 
\textbf{Stability.} The most interesting part of our experiment is that PUT strengthens the stability in adversarial learning. As depicted in Fig.~\ref{fig:Stable}, `Baseline' (i.e., CUT) gives a unstable convergence curve despite CL is used. We use top-3, top-5 and top-25 negative samples in RankNCE for comparison. To our surprise, `PUT-3' and `PUT-5' achieve better stability and superior performance by using fewer negative samples. 

\textbf{Efficiency.} As shown in Tab.~\ref{tab:my-table}, we compare the average running time of different methods for reconstructing all images in the Horse$\rightarrow$Zebra dataset. Compared with CUT, the proposed achieves superior results with similar efficiency. Compared with NEGCUT and DCLGan, the proposed model demonstrates better efficiency and significant performance improvements. Though our model reaches a similar level of efficiency among state-of-the-art methods, we achieve solid performance and help to stabilize the training process. The above results show that PUT outperforms the existing models while promising competitive efficiency.

\begin{table*}[t]
\centering
\begin{tabular}{cccccccc}
\hline
\multirow{2}{*}{Method} &

\multicolumn{4}{c}{CityScapes}      & \multicolumn{3}{c}{Horse$\rightarrow $ Zebra} \\ \cmidrule(lr){2-5} \cmidrule(lr){6-8} 
                        & mAP {\color{red}$\uparrow$}   & pixAcc{\color{red}$\uparrow$} & classAcc{\color{red}$\uparrow$} & FID {\color{red}$\downarrow$}   & FID{\color{red}$\downarrow$}      & Speed(s)   & Memory     \\ \hline
CUT                     & 22.29 & 75.22 & 29.6    & 62.19  & 45.51      & 0.245      & 3.053   \\
Fast CUT                & 18.05 & 66.63 & 24.2    & 99.18 &  73.37    & \textcolor{blue}{0.150}       & \textcolor{red}{2.573}   \\
FSeSim                  & 19.06 & 66.00   & 26.7    & 51.23 & 43.80     & \textcolor{red}{0.107}      & \textcolor{blue}{2.920}    \\
DCLGan                  & 22.91 & 76.97 & 29.6    & 47.83  & 42.64    & 0.389      & 7.014   \\
NEGCUT                  & 22.83 & 77.04 & 29.0    & \textcolor{blue}{42.65} & 39.63    & 0.571      & 4.754   \\ \hline 
\textbf{PUT-3}                   & \textcolor{blue}{24.31} & \textcolor{blue}{78.28} & \textcolor{red}{31.7}    & \textcolor{red}{41.43} & \textcolor{blue}{33.82}    & 0.245      & 3.053   \\
\textbf{PUT-5}                   &\textcolor{red}{24.86} & \textcolor{red}{79.17} &\textcolor{blue}{ 30.1}    & 42.99 & \textcolor{red}{ 33.62}    & 0.247      & 3.053   \\ \hline
\end{tabular}
\caption{Quantitative Comparison. Our method outperforms state-of-the-art methods across all evaluation metrics.}
\vspace{-3mm}
\label{tab:my-table}
\end{table*}

\begin{table*}[h]
\begin{tabular}{ccccccc}
\hline
\multirow{2}{*}{method} & Training setting & \multicolumn{4}{c}{CityScapes}       & Horse$\rightarrow$Zebra \\ 
\cmidrule(lr){2-2}\cmidrule(lr){3-6}\cmidrule(lr){7-7}
                        & layer           & FID{\color{red}$\downarrow$}     & pixAcc{\color{red}$\uparrow$} & classAcc{\color{red}$\uparrow$} & mAp{\color{red}$\uparrow$}    & FID{\color{red}$\downarrow$}         \\ \hline
w/ solo RankNCE         & 1             & 42.76 & 30.81 & 76.24   & 24.01 & 41.33      \\ \hline
w/ Multi-layer RankNCE   & 5              & \textcolor{red}{41.43} & \textcolor{red}{31.71} & \textcolor{red}{78.28}   & \textcolor{red}{24.31} & \textcolor{red}{33.82}       \\ \hline
\end{tabular}
\caption{Ablation on Multi-layer RankNCE.}
\vspace{-4mm}
\label{tab:ABres}
\end{table*}

\subsection{Ablations} We perform analyses to verify the effectiveness of each component in PUT, including RankNCE loss, and multi-layer RankNCE. To analyze RankNCE, we use the `pruning' and `ranking' sub-modules separately. In the Fig.~\ref{fig:ablation_put}, `w/ pruning' show better stability with similar FID result. Though the training curve becomes more stable, a solo ranking/pruning strategy is fail to improve image quality. Then, we include ranking and pruning operations, and observe that the training curve becomes more stable and the FID is consistently improved.

We conduct an ablation study of the multi-layer RankNCE, and analyze the experimental results on Horse$\rightarrow$Zebra and CityScapes. As depicted in Tab.~\ref{tab:ABres}, `w/ Multi-layer RankNCE' exhibits the best results on the FID index, which reveals that multi-layer RankNCE is effective for I2I translation.

\subsection{User study} To further demonstrate the quality of our results, we conduct a user study according to the mean opinion score. We select 50 images randomly and forward them with baseline models to obtain the results for comparison. We recruited 10 volunteers for testing and asked volunteers to rank the results based on perception quality. As shown in Fig.~\ref{fig:ABtest}, our method shows advantages over other methods by achieving 57\% Top-1 preference.

\subsection{High-Resolution Effect}
To further evaluate the versatility of our algorithm, we conduct PUT for an experiment of the single image high-resolution translation, which means the source/target domain only has one picture and they are totally unpaired. In this task, we transfer Claude Monet’s paintings to reference natural photographs by following prior works~\cite{zheng2021spatially,park2020contrastive}. Specifically, the generators and discriminators of our model are based on StyleGAN2~\cite{karras2020analyzing}, and we use gradient penalty to stabilize optimization~\cite{mescheder2018training}. In particular, the RankNCE loss is used for identity-preserving.

\textbf{Result.}We compare our model with baseline methods including neural style transformation methods (Gatys et al~\cite{gatys2016image}) and the latest single image translation methods (CUT~\cite{park2020contrastive} and FSeSim~\cite{zheng2021spatially}) for a qualitative comparison. As shown in Fig.~\ref{fig:hr_effect}, our model produces a higher quality result. For example, our model successfully captures the style of the target image while preserving the structure of the input well.

\section{Conclusion and Limitations}
\label{sec:conclusion}

In this paper, we propose a novel model called PUT for unpaired image-to-image translation. Compared with the previous contrastive learning methods, our proposed PUT is stable to learn the information between the corresponding patches, leading to a more effective contrast learning system. 

However, there still exist some limitations. For instance, current image translation methods, including PUT, could fail when the given category, which is recognized as source instance, has too few pixels in the input image. How to effectively address these problems still remains an open question, which we will discuss in our future work.

\bibliographystyle{ACM-Reference-Format}
\balance
\bibliography{egbib}
\end{document}